\documentclass[11pt,a4paper]{article}
\usepackage[hyperref]{naaclhlt2018}
\usepackage{times}
\usepackage{latexsym}
\usepackage{slashbox}

\usepackage[english]{babel}
\usepackage[utf8]{inputenc}
\usepackage[T1]{fontenc}
\usepackage{amsfonts} 
\usepackage{booktabs}
\usepackage{setspace}
\usepackage{subcaption}
\usepackage{amsmath}
\usepackage{graphicx}
\usepackage[colorinlistoftodos,prependcaption,textsize=tiny]{todonotes}

\newcommand{\stasEE}[1]{\textcolor{black}{#1}}

\aclfinalcopy

\title{Commonsense mining as knowledge base completion? A study on the impact of novelty}

\author{Stanisław Jastrzębski\thanks{\indent Work partially done as intern at MILA}\\
Jagiellonian University \\
{\small \texttt{stanislaw.jastrzebski@uj.edu.pl}}
\And
Dzmitry Bahdanau  \\
MILA\\
Université de Montréal
\And
Seyedarian Hosseini  \\
MILA\\
Université de Montréal
\AND
 Michael Noukhovitch \\
MILA\\
Université de Montréal
\And
Yoshua Bengio\thanks{\indent CIFAR Senior Fellow} \\
MILA\\
Université de Montréal \\
\And
Jackie Chi Kit Cheung \\
MILA\\
McGill University
}

\date{}

\begin{document}
\maketitle

\begin{abstract}
Commonsense knowledge bases such as ConceptNet represent knowledge in the form of relational triples. Inspired by the recent work by~\newcite{P16-1137}, we analyse if knowledge base completion models can be used to mine commonsense knowledge from raw text. We propose \emph{novelty} of predicted triples with respect to the training set as an important factor in interpreting results. We critically analyse the difficulty of mining novel commonsense knowledge, and show that a simple baseline method outperforms the previous state of the art on predicting more novel triples.
\end{abstract}

\section{Introduction}

Many natural language understanding tasks require commonsense knowledge in order to resolve ambiguities involving implicit assumptions. Collecting such knowledge and representing it in a reusable way is thus an important challenge. There exist several commonsense knowledge bases maintained by experts (CyC) or acquired by crowdsourcing (ConceptNet) which represent commonsense knowledge as relational triples, e.g. \textit{(``pen'', ``UsedFor'', ``writing'')}~\citep{Liu:2004:CMP:1031314.1031373}. Automatic mining of commonsense knowledge, the focus of this work, aims to improve the coverage of such resources.

One common way of improving the coverage of knowledge bases is through knowledge base completion (KBC), which can be formalized as predicting the existence of edges between (usually) pre-existing nodes in the graph. Recent work by~\newcite{P16-1137} approached commonsense mining as a KBC task. Their method mines candidate triples from Wikipedia and reranks the triples with a KBC model in order to extend ConceptNet.

The goal of this paper is to investigate why recent systems such as the above achieve good performance, and understand their potential for mining commonsense. We approach it by breaking down the previously reported aggregate results into the cases in which models perform well or poorly. We focus in particular on the issue of the \emph{novelty} of model predictions with respect to the triples in the training set. For example, a triple predicted by a system could be correct because it generates output with a slightly different wording or morphological inflection, e.g. \textit{(``fish'', ``AtLocation'', ``water'')} from \textit{(``fish'', ``AtLocation'', ``in water'')}, or it could be correct because it exhibits some degree of semantic generalization, e.g. \textit{(``fish'', ``IsCapableOf'', ``swimming'')} from \textit{(``fish'', ``AtLocation'', ``in water'')}. Arguably, the former could be handled by better standardization of data set formats or more comprehensive model pre-processing, whereas the latter presents an example of genuine commonsense inference and novelty. This analysis is especially important for commonsense mining because of the diversity of the entities, relations, and linguistic expressions thereof in current datasets. 

The contribution of this paper is two-fold. First, we test if the KBC task as it is set up in recent work can gauge a model's ability to mine novel commonsense (i.e. find novel commonsense facts based on some resource). We observe the contrary. We present a model that performs poorly on KBC but matches the best model on the task of \stasEE{mining novel commonsense (evaluated by re-ranking extracted candidate triples from Wikipedia)}. We then examine the cause of this discrepancy, and find that around $60\%$ of triples in the KBC test set used by~\newcite{P16-1137} are minor rewordings of existing triples in the training set. This suggests that controlling for the novelty of triples in both KBC and Wikipedia evaluation is needed.

Second, we present a reassessment of previous methods in which we control the dataset for novelty, extending the results of~\newcite{P16-1137}. We introduce a simple automated novelty metric and show that it correlates with human judgment. We then show that the performance of most models on both KBC and Wikipedia triple reranking drops drastically when we evaluate them on examples that are genuinely new according to our metric. Finally, we demonstrate that a simple baseline model that does not model all interactions between elements in a triple performs surprisingly well on both KBC and reranking when we focus on novel triples.

\section{Related work}

Knowledge extraction from text corpora is a vast research area \citep{banko_open_2007, mitchell2015}, yet works that specifically target commonsense knowledge are comparatively rare \citep{gordon_inferential_2014}. Our focus is on the specific approach to mining commonsense knowledge by casting it as a KBC task, as in~\newcite{P16-1137} and ~\newcite{DBLP:conf/acl/ForbesC17}.

Knowledge base completion (KBC) is a method to improve coverage of knowledge base by predicting non-existing edges between nodes ~\citep{nickel2011three, socher_reasoning_2013_fix}. \stasEE{A common modeling approach to KBC is to embed both nodes and the edge into a common representation space, followed by a simple prediction model~\citep{socher_reasoning_2013_fix}.}

Recently, \newcite{DBLP:journals/corr/DettmersMSR17} observed that some KBC benchmarks have test set triples that are simply inversions of triples in their training sets. Our work draws attention to a related issue in commonsense KBC. Additionally, we find that simple baseline models achieve strong performances in our setting, in agreement with other studies of KBC ~\citep{DBLP:journals/corr/abs-1710-10881, DBLP:conf/rep4nlp/KadlecBK17}.

In \newcite{angeli2013philosophers}, triple retrieval based on distributional similarity is used to complete ConceptNet. Our procedure for determining the novelty of the triple is similar, but we apply it only in the context of evaluation.

\section{Completion vs Mining}
\label{sec:completion_vs_mining}

\stasEE{Our goal in this section is to analyse the relation between KBC and commonsense mining tasks following the setup of~\newcite{P16-1137}.}

\subsection{Models}
All our models take $(h, r, t)$ triples as inputs, where $h$ and $t$ are sequences of words representing concepts and $r$ is a relation from the ConceptNet schema, \stasEE{and output the probability that the triple is true}.  Following \newcite{P16-1137}, we embed $h$ and $t$ by computing the sums $\mathbf{h}$ and $\mathbf{t}$ of the respective word vectors. 

\newcite{levy2015supervised} showed that, in the context of predicting the hypernymy relation, using only head or only tail can be a strong baseline. To better understand how complex reasoning is needed for both KBC and mining tasks, we similarly consider the two following models, which make strong simplifying assumptions about the dependencies between elements in a triple. The \textbf{Factorized} model uses only two-way interactions to compute the triple score: 
\begin{equation}
\begin{aligned}
s(\mathbf{h}, \mathbf{r}, \mathbf{t}) &= \alpha \langle \mathbf{A} \mathbf{h} + \mathbf{b_1},  \mathbf{B} \mathbf{t} + \mathbf{b_2} \rangle \\
& + \beta \langle \mathbf{A} \mathbf{r} + \mathbf{b_1},  \mathbf{B} \mathbf{t} + \mathbf{b_2} \rangle \\
& + \gamma \langle \mathbf{A} \mathbf{r} + \mathbf{b_1},  \mathbf{B} \mathbf{h} + \mathbf{b_2} \rangle, \label{eq:factorized} \\
\end{aligned}
\end{equation}
\stasEE{where $\mathbf{h}$, $\mathbf{r}$, $\mathbf{t}$ are $d_1$ dimensional embeddings of head, relation and tail, $\mathbf{A}, \mathbf{B}$ are $d_1 \times d_2$ matrices, $\mathbf{b_1}, \mathbf{b_2}$ are $d_2$ dimensional biases, and $\alpha$, $\beta$, $\gamma$ are learned scalars. The \textbf{Prototypical} model is similar, but considers only the head-to-relation and tail-to-relation terms (first and third terms in Eq.~\ref{eq:factorized}}). 

\stasEE{We compare the two new models with the best model  from~\newcite{P16-1137}, a single hidden layer \textbf{DNN}. In that model, the triple score is computed as} 
\begin{equation}
\begin{aligned}
u(\mathbf{h}, \mathbf{t}) = \phi (\mathbf{A}\mathbf{h} + \mathbf{B}\mathbf{t} + \mathbf{b_1})\\
s(\mathbf{h}, \mathbf{r}, \mathbf{t}) =  \mathbf{W} u(\mathbf{h}, \mathbf{t}) + \mathbf{b_2},
\end{aligned}
\end{equation}
\stasEE{where $\phi$ is a nonlinearity, $\mathbf{A}$, $\mathbf{B}$ are $d_1 \times d_2$ matrices, $\mathbf{b_1}$ is a $d_2$ dimensional bias, $W$ is a $d_2$ dimensional vector and $\mathbf{b_2}$ is a scalar bias. Additionally, we compare against the \textbf{Bilinear} model of~\newcite{P16-1137}\footnote{It is the only model evaluated against the Wikipedia ranking task in ~\newcite{P16-1137}.}, which computes the triple score as}
\begin{equation}
\begin{aligned}
s(\mathbf{h}, r, \mathbf{t}) =  \mathbf{h^T} \mathbf{M}_r  \mathbf{t} ,
\end{aligned}
\end{equation}
\stasEE{where $\mathbf{M}_r$ is a $d_1 \times d_1$ dimensional matrix, separate for each relation in the dataset.} 

All models' scores are fed into a sigmoid function in order to compute the final prediction.

\subsection{Setup}

KBC models are trained using $100,000$ triples from ConceptNet5 that were extracted from the Open Mind Common Sense (OMCS) corpus~\citep{SPEER12.1072}. For evaluation, \stasEE{we consider two ways to split the dataset}: a random split, as well as the confidence-based split proposed by \citep{P16-1137}, which uses triples with the highest ConceptNet confidence scores as a test set.\footnote{We note that the random test set consists of worse-quality triples than the confidence-based split. However, the latter leads to a serious bias in evaluation. We leave addressing this trade-off for future work.} Following ~\newcite{P16-1137}, negative examples are sampled by randomly swapping the head, tail, or relation component of each triple with a different head, tail, or relation in the dataset". The cross-entropy loss is used, and \stasEE{models are evaluated using F1 score.\footnote{The threshold is selected based on a separate development set, as in~\newcite{SPEER12.1072}. }} All models are initialized using skip-gram embeddings that were pretrained on the OMCS corpus. 

\stasEE{The commonsense mining task is based on a set of $1.7$ M extracted candidate triples from Wikipedia by \citet{P16-1137}. The extracted triples are ranked using a KBC model, and the top of the ranking is manually evaluated.} We will refer to the experiments in which we rerank external candidate triples as \textit{mining} experiments.

We found that similar hyperparameters and optimization methods work well across the models. We use $1,000$ hidden units, and apply L2 regularization with a weight of $10^{-6}$ to the word embeddings. All models are optimized using Adagrad~\citep{Duchi:EECS-2010-24} with a learning rate of $0.01$ and batch sizes of $200$ (DNN) and $600$ (Factorized and Prototypical). In Section~\ref{sec:order}, we compare against the scores of a Bilinear model provided by~\newcite{P16-1137}. Experiments are performed using Keras~\citep{chollet2015keras} and TensorFlow~\citep{tensorflow2015-whitepaper}.

\subsection{Comparison of KBC and Wikipedia evaluations}
\label{sec:order}

\stasEE{First, we directly test if the performance of a model on the KBC task is predictive of its performance on the mining task}. We follow the mining evaluation protocol from~\citep{P16-1137}: we rank triples by assigned scores and manually evaluate the top $100$ resulting triples on a scale from $0$ (nonsensical) to $4$ (true statement). We re-evaluate their model against our baselines and find that the knowledge base completion task is a poor indicator of performance on Wikipedia. Even though the Factorized and Prototypical models achieve a similar or worse score compared to DNN on the KBC task (see the first row of Table~\ref{tab:results_confident}), their mining performance on the top $100$ triples is better than both DNN and the Bilinear model (see Table~\ref{tab:reeval}). Triples were scored by two students and scores were averaged, with $0.81$ Pearson correlation and $0.48$ kappa inter-annotator agreement.

\begin{table}
\small

\begin{tabular}{llll}
\toprule
{\backslashbox{Novelty}{Model}} &     DNN & Factorized & Prototypical \\
\midrule
Entire       &   $\mathbf{0.892}$ & $0.890$  & $0.794$  \\
$\leq 33\%$   &$\mathbf{0.950}$ &  $0.922$ & $0.911$ \\
$(33\%, 66\%]$ &$\mathbf{0.920}$ & $0.898$ & $0.839$  \\
 $\ge 66\%$ & $0.720$ & $\mathbf{0.821}$ & $0.574$ \\
\bottomrule
\end{tabular}

\caption{\stasEE{F1 scores} on \newcite{P16-1137} confidence-based test set. F1 score is reported on each bucket \stasEE{(based on the percentile of triple novelty)} and the entire test set. } 
\label{tab:results_confident}
\end{table}

\begin{table}
\small
\begin{tabular}{lllll}
\toprule
{} & Bilinear & Factorized & Prototypical & DNN \\
\midrule
Wikipedia &     2.04 &       \textbf{2.61} &         2.55 & 2.5 \\
\bottomrule
\end{tabular}
\caption{\stasEE{Average human-assigned score (from $1$ to $5$) of the top $100$ Wikipedia triples ranked  by baselines compared to DNN and Bilinear from ~\citet{P16-1137}.}}
\label{tab:reeval}
\end{table}

\subsection{Novelty of triples}
\label{sec:trivial}

We hypothesize that the discrepancy reported in Section \ref{sec:order} is due to a strong overlap of the training and testing sets in the KBC setup of \citet{P16-1137}. We perform a human evaluation of the novelty of the triples in the three test sets with respect to the $100,000$ triples in the ConceptNet training set. The first is the confidence-based test set used in \newcite{P16-1137}. We compare it with a random subset of ConceptNet. Finally, we consider a sample of $300$ triples from the top $10,000$ triples of the Wikipedia dataset ordered by the Bilinear model.

For each triple in the three datasets, we fetch the five closest neighbours using word embedding distance and divide them into five categories based on the closest triple found in the training set: \textit{``same relation and minor rewording''} ($1$), \textit{``different relation and minor rewording''} ($2$), \textit{``same relation and related word''} ($3$), \textit{``different relation and related word''} ($4$), \textit{``no directly related triple''} ($5$). We ignore a small percentage of triples that are not describing commonsense knowledge, as well as false triples (some in the random subset, and a large percentage in the Wikipedia dataset). 

\stasEE{To give a better intuition, we provide example triples for the confidence-based split of \newcite{P16-1137}. In Category 1 (defined as ``same relation and minor rewording''), we find \textit{(``egg'', ``IsA'', ``food'')}, which has a close analog in the training set: \textit{(``egg'', ``IsA'', ``type of food'')}. An example of a test triple in Category 3 (defined as \textit{``different relation and related word''}) is \textit{(``floor'', ``UsedFor'', ``walk on'')}, which has a corresponding triple in the training set  \textit{(``floor'', ``UsedFor'', ``stand on'')}. In the Appendix, we provide more examples of triples from each category.}

As shown in Table~\ref{tab:nov2}, we observe that approximately $87\%$ of examples in the \stasEE{confidence-based} test set fall into the first or second category, while these categories constitute only $19\%$ of the considered subset of the Wikipedia triples (even after filtering out false triples). We argue that not controlling for the novelty of triples might introduce hard-to-predict biases in the evaluation.

%surprisingly

Finally, to understand the effects of using the confidence-based split, we also re-evaluate models on a random split. We observe that scores are consistently lower than on the confidence-based split (compare the first rows of Tables~\ref{tab:results_confident} and \ref{tab:results_random}). \stasEE{Interestingly, the overall performance of the DNN model degrades the most (with an absolute difference in F1 score of $9\%$), compared to Prototypical ($4\%$) and Factorized ($7\%$).}

\section{Evaluation using novelty metric}

\stasEE{Motivated by the described similarity of the train and test sets in the KBC task, we shift our attention to re-evaluating models on datasets controlled for novelty, extending results of~\citet{P16-1137}. We consider the same tasks as in Sec.~\ref{sec:completion_vs_mining}: the ConceptNet5 completion task and the commonsense mining task based on Wikipedia triples.}

\subsection{Automatically measuring novelty}
 
 To approximate novelty, we use word embeddings (computed over the OMCS corpus) to calculate distance $d(a, b) = ||head(a) - head(b)||_2 + ||tail(a) - tail(b)||_2$, where head and tail are represented by the average of word embeddings. Such a formulation is related to the concept of \emph{paradigmatic} similarity~\citep{sahlgren06}, and word embedding-based distance can approximate paradigmatic similarity~\citep{P15-1014}. Two words are paradigmatically similar if one can be replaced by the other, while maintaining syntactical correctness of the sentence (e.g. \textit{``The wolf/tiger is a fierce animal''}). We observe that many trivial test triples are characterized by the existence of a triple in the training set that only differs by such substitutions.

We observe that the proposed distance metric is correlated with human-assigned novelty scores \stasEE{(from Sec.~\ref{sec:trivial})}. On the considered datasets, the Pearson correlation between the automatic novelty score and the human-assigned novelty score is $0.22$ to $0.47$, with p-values between $0.03$ and $0.004$. We acknowledge that the automated metric is simplistic; for instance, it underperforms for the triples containing rare words or long phrases. Nevertheless, the metric enables the detection of a substantial portion of trivial triples (e.g. morphological variations), and we leave developing better measures of novelty for future work.

\begin{figure}
\centering
\includegraphics[width=0.31\textwidth]{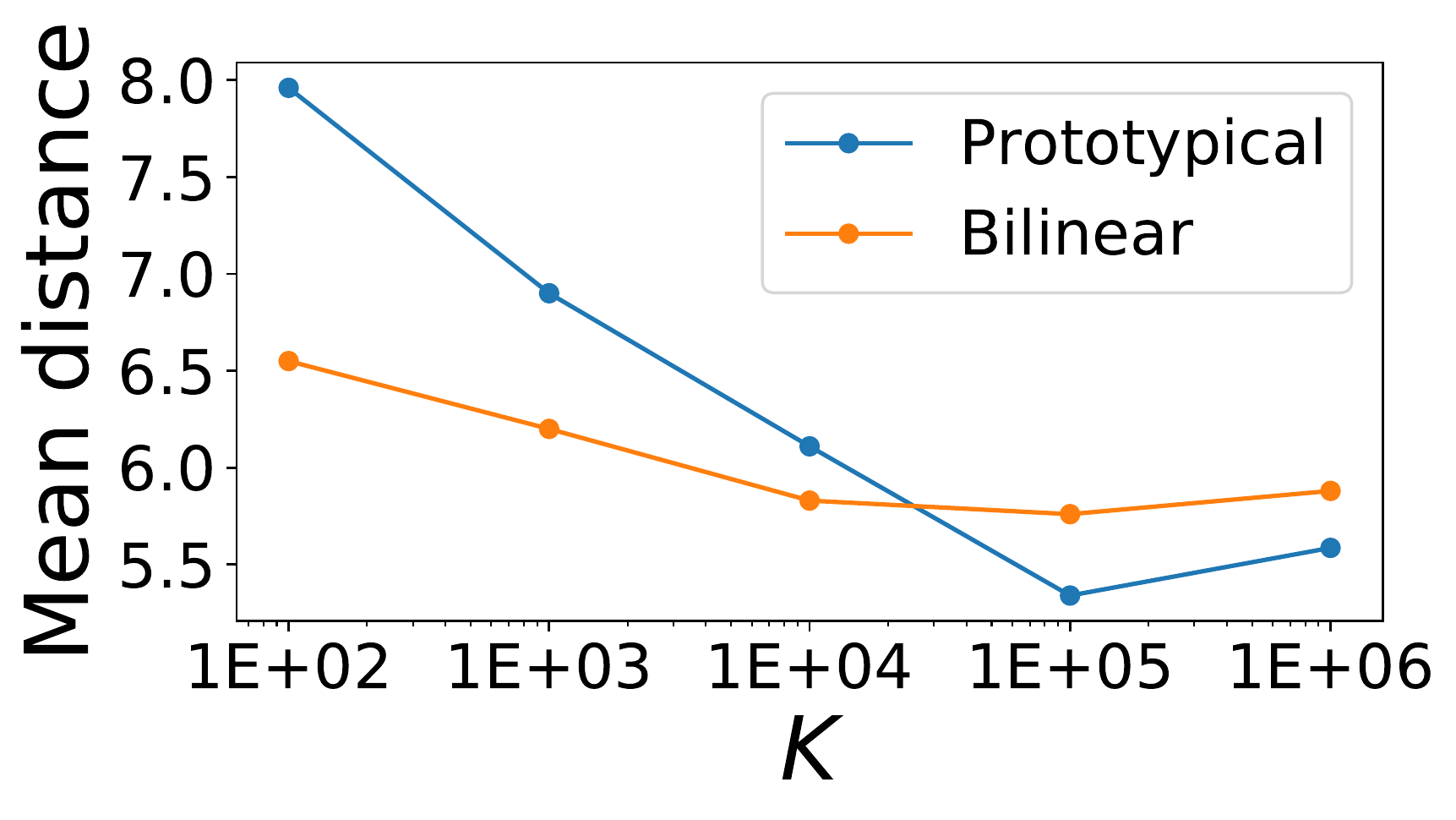}
\caption{Mean \stasEE{embedding} distance (y axis) of top $K$ (x axis) of triples in Wikipedia dataset for Bilinear (orange) and Prototypical (blue).}
\label{fig:top_k_dist}
\end{figure}

Using the introduced metric, we can partially explain the inconsistency in the performance of the Prototypical and Bilinear models between KBC and mining Wikipedia. We note that the top of the ranking on Wikipedia consists of mostly very far (novel) triples (Figure~\ref{fig:top_k_dist}), while the KBC confidence-based test set is mostly composed of trivial triples (as argued in Section ~\ref{sec:trivial}).

\begin{table}
\centering
\small
\begin{tabular}{llllll}
\toprule
{\backslashbox{Dataset}{Novelty}} &      1 &     2 &      3 &     4 &      5 \\
\midrule
Wikipedia         &  $14$\% &   $5$\% &  $17$\% &  $8$\% &   $44$\% \\
Confident &  $65$\% &  $22$\% &   $4$\% &  $4$\% &   $2$\% \\
Random       &  $21$\% &  $10$\% &  $16$\% &  $3$\% &  $29$\% \\
\bottomrule
\end{tabular}
\caption{Human-assigned novelty \stasEE{categories} to triples from $3$ different test datasets. High-quality triples are usually trivial. The columns report the percentage of triples in each category ordered by novelty. \stasEE{Category 1 corresponds to \textit{``same relation and minor rewording''}. Category 5 corresponds to \textit{``no directly related triple''}.} }
\label{tab:nov2}
\end{table}

\subsection{Novelty-binned evaluation of KBC}
\label{sec:novkbc}

We now re-evaluate the \stasEE{KBC} models using our proposed novelty metric. First, we examine the performance on different subsets of the confidence-based split \stasEE{of ConceptNet5}. Specifically, we split the confidence-based test set into $3$ buckets, according to $33\%$ ($1.93$ distance) and $66\%$ ($2.80$ distance) quantile of distance to the training set. Second, we run a similar experiment but on a random split of the training set (bucket thresholds at $2.1$ and $2.95$). Results are reported in Tables~\ref{tab:results_confident} and~\ref{tab:results_random}. 

As expected, the performance of models degrades quickly across buckets. The F1 score of the farthest bucket is 10 to 20\% lower than the F1 score of the closest bucket. We observe that the Factorized model achieves the strongest performance on the farthest bucket.

\begin{table}
\small
\centering

\begin{tabular}{llll}
\toprule
{\backslashbox{Novelty}{Model}} &     DNN & Factorized & Prototypical \\
\midrule
Entire       &   $0.809$ & $\mathbf{0.822}$  & $0.755$  \\
$\leq 33\%$   &$\mathbf{0.883}$ &  $0.874$ & $0.866$  \\
$(33\%, 66\%]$ &$0.809$ & $\mathbf{0.812}$ & $0.758$  \\
 $\ge 66\%$ & $0.725$ & $\mathbf{0.731}$ & $0.674$ \\
\bottomrule
\end{tabular}

\caption{F1 scores on random split. F1 score is reported on each bucket \stasEE{(based on the percentile of triple novelty)} and the entire set.}
\label{tab:results_random}
\end{table}

\subsection{Novelty-binned evaluation on Wikipedia}

Similar to Section~\ref{sec:novkbc}, we analyse splitting candidate triples \stasEE{for the mining task} using our novelty metric. We split the Wikipedia dataset into $3$ buckets based on $33\%$ ($3.21$ distance) and $66\%$ ($4.22$ distance) quantiles of distance to the training set, and we manually score the top $100$ triples in each bucket \stasEE{on the same scale from $1$ to $5$.}

As in Section~\ref{sec:novkbc}, we note a degradation of performance across buckets for all models (from $1.06$ to $0.32$ mean human-assigned score) and again the Factorized model achieves the best performance on the farthest bucket (mean score $2.26$ compared to $1.63$ and $1.41$). The Factorized model outperforms DNN on all buckets despite being a simpler model, which we hypothesize is due to DNN being more prone to overfitting.

\begin{table}
\small
\centering
\begin{tabular}{llll}
\toprule
{\backslashbox{Novelty}{Model}} &     DNN & Factorized & Prototypical \\
\midrule
$\leq 33\%$   &$2.47$ &  $\mathbf{2.58}$ & $2.33$  \\
$(33\%, 66\%]$ &$2.34$ & $\mathbf{2.41}$ & $2.24$  \\
 $\ge 66\%$ & $1.41$ & $\mathbf{2.26}$ & $1.63$ \\
\bottomrule
\end{tabular}

\caption{Novelty-based evaluation of quality of mined triples from Wikipedia dataset. \stasEE{Triples are scored by humans on a scale from $1$ to $5$.}}
\label{tab:novelty}
\end{table}

\section{Conclusions}

Mining genuinely novel commonsense is a challenging task, and training successful models will require large training sets (e.g. ConceptNet) and principled evaluation. We critically assess the potential of KBC models for mining commonsense knowledge, and propose several first steps towards a more principled evaluation methodology. Future work could focus on developing better novelty metrics, and developing new regularization techniques to better generalize to novel triples.

\section*{Acknowledgments}

SJ was supported by Grant No.~DI 2014/016644 from Ministry of Science and Higher Education, Poland. Work at MILA was funded by NSERC, CIFAR, and Canada Research Chairs.

\bibliographystyle{acl_natbib}
\bibliography{ref,dimazotero}

\appendix

\section{Example triples}

In this Appendix we report randomly picked examples from the human-assigned novelty categories considered in the paper for each of the $3$ datasets. Due to the large size of the training set, instead of showing all of the triples from the training set to the human annotators, we show only the $5$ closest triples using the embedding-based distance. A triple is classified as belonging to the given category if \emph{at least one} of the retrieved triples is sufficiently related. For example, if for \textit{(``egg'', ``IsA'', ``food'')} we find the triple \textit{(``egg'', ``IsA'', ``type of food'')} in the top $5$ closest examples, we categorize it as belonging to the first category (``same rel, rephrase'').

\subsection{Confidence-based split}In this Section we report examples for each novelty category from the confidence-based split dataset. For each example we include the $5$ triples that were shown to the human annotator, ordered by closeness according to our word embedding metric.\subsubsection{``same rel, rephrase''}
        \begin{itemize}
        
  \item (egg, IsA, food): \textit{\,(egg, UsedFor, food),}\textit{\,(egg, HasProperty, good for food),}\textit{\,(egg, IsA, type of food),}\textit{\,(egg, HasProperty, good for you),}\textit{\,(egg, AtLocation, omletts),}
  \item (book, AtLocation, classroom): \textit{\,(lot book, AtLocation, classroom),}\textit{\,(physic, AtLocation, classroom),}\textit{\,(teacher aide, AtLocation, classroom),}\textit{\,(desk and chair, AtLocation, classroom),}\textit{\,(test paper, AtLocation, classroom),}
  \item (dog, CapableOf, be pet): \textit{\,(dog, CapableOf, be great pet),}\textit{\,(dog, CapableOf, be loyal pet),}\textit{\,(dog, CapableOf, be over-fed),}\textit{\,(dog, IsA, good pet),}\textit{\,(dog, NotDesires, be with cat)}
 \end{itemize}\subsubsection{``different rel, rephrase''}
        \begin{itemize}
        
  \item (window, MadeOf, glass): \textit{\,(window, HasProperty, make of glass),}\textit{\,(window, DefinedAs, glass that be stick to window frame),}\textit{\,(abottle, MadeOf, glass),}\textit{\,(window, UsedFor, look out of),}\textit{\,(window, UsedFor, look inside),}
  \item (bury cat, HasSubevent, dig hole): \textit{\,(bury cat, HasFirstSubevent, dig hole),}\textit{\,(bury cat, HasSubevent, dig),}\textit{\,(bury cat, HasFirstSubevent, dig grind),}\textit{\,(bury cat, UsedFor, when your cat be dead),}\textit{\,(bury cat, HasPrerequisite, make sure it be dead),}
  \item (bridge, UsedFor, cross river): \textit{\,(bridge, CapableOf, cross river),}\textit{\,(bridge, UsedFor, cross sometihng),}\textit{\,(bridge, UsedFor, cross water),}\textit{\,(bridge, UsedFor, cross over),}\textit{\,(bridge, ReceivesAction, find over river)}
 \end{itemize}\subsubsection{``same rel, similar word''}
        \begin{itemize}
        
  \item (cat, CapableOf, hunt mouse): \textit{\,(cat, CapableOf, hunt lizard),}\textit{\,(cat, NotCapableOf, like mouse),}\textit{\,(cat, UsedFor, kill mouse),}\textit{\,(cat, CapableOf, kill mouse),}\textit{\,(cat, Desires, eat mouse),}
  \item (pilot, CapableOf, land airplane): \textit{\,(pilot, CapableOf, carsh airplane),}\textit{\,(pilot, CapableOf, land taildragger),}\textit{\,(pilot, CapableOf, work in airplane),}\textit{\,(pilot, CapableOf, land),}\textit{\,(pilot, AtLocation, airplane),}
  \item (play sport, HasSubevent, run): \textit{\,(play baseball, HasSubevent, run),}\textit{\,(play frisbee, Causes, run),}\textit{\,(do some exercise, HasSubevent, run),}\textit{\,(horse jump high when they, HasProperty, run),}\textit{\,(go for run, HasSubevent, run)}
 \end{itemize}\subsubsection{``different rel, similar word''}
        \begin{itemize}
        
  \item (statue, AtLocation, museum): \textit{\,(statue, ReceivesAction, see in museum),}\textit{\,(statue, IsA, example of art),}\textit{\,(statue, UsedFor, imortalize someone),}\textit{\,(statue, HasProperty, hard to create),}\textit{\,(statue, CapableOf, be beautiful),}
  \item (son, PartOf, family): \textit{\,(son, IsA, member of family),}\textit{\,(man and his daughter, IsA, family),}\textit{\,(son, DefinedAs, child of parent),}\textit{\,(son, AtLocation, his home),}\textit{\,(son, IsA, male kid of his parent),}
  \item (internet, UsedFor, research): \textit{\,(internet, IsA, amaze research tool),}\textit{\,(go on internet, UsedFor, research),}\textit{\,(internet, IsA, research project of darpa),}\textit{\,(internet, UsedFor, do research or chat),}\textit{\,(internet, HasA, lot of information)}
 \end{itemize}\subsubsection{``no directly related triple''}
        \begin{itemize}
        
  \item (clerk, CapableOf, stock shelve): \textit{\,(clerk, CapableOf, be bag grocery),}\textit{\,(clerk, CapableOf, price item),}\textit{\,(clerk, CapableOf, bag grocery),}\textit{\,(clerk, CapableOf, enter data),}\textit{\,(clerk, AtLocation, at hotel),}
  \item (human, HasA, five finger on each hand): \textit{\,(human, HasA, five toe on each foot),}\textit{\,(human, HasA, arm hand finger fingernail and lunula),}\textit{\,(human, HasA, two hand),}\textit{\,(human, CapableOf, write with right hand),}\textit{\,(human, CapableOf, stand on two leg),}
  \item (cat, CapableOf, corner mouse): \textit{\,(cat, NotCapableOf, like mouse),}\textit{\,(cat, CapableOf, kill mouse),}\textit{\,(cat, UsedFor, kill mouse),}\textit{\,(cat, UsedFor, keep mouse away),}\textit{\,(cat, AtLocation, petstore)}
 \end{itemize}
\subsection{Random split}In this Section we report examples for each novelty category from Random split dataset. For each example we include the $5$ triples that were shown to the human annotator, ordered by closeness according to our word embedding metric.\subsubsection{``same rel, rephrase''}
        \begin{itemize}
        
  \item (coffee mug, AtLocation, cupboard): \textit{\,(mug, AtLocation, cupboard),}\textit{\,(coffee cup, AtLocation, cupboard),}\textit{\,(tea cup, AtLocation, cupboard),}\textit{\,(cup and plate, AtLocation, cupboard),}\textit{\,(can of soup, AtLocation, cupboard),}
  \item (man, IsA, person): \textit{\,(man, IsA, male person),}\textit{\,(egoistic person, IsA, person),}\textit{\,(woman, IsA, person),}\textit{\,(child, InheritsFrom, person),}\textit{\,(child, IsA, person),}
  \item (bookshelf, IsA, for store book): \textit{\,(bookshelf, UsedFor, store book),}\textit{\,(bookshelf, UsedFor, display and store read material),}\textit{\,(bookshelf, UsedFor, hold and organize book),}\textit{\,(bookshelf, UsedFor, organize book),}\textit{\,(bookshelf, UsedFor, display book)}
 \end{itemize}\subsubsection{``different rel, rephrase''}
        \begin{itemize}
        
  \item (hear sing, HasSubevent, listen): \textit{\,(hear sing, HasFirstSubevent, listen),}\textit{\,(hear sing, HasPrerequisite, listen),}\textit{\,(hear, HasPrerequisite, listen),}\textit{\,(hear music, HasPrerequisite, listen),}\textit{\,(hear music, HasSubevent, listen),}
  \item (procreate, HasPrerequisite, find mate): \textit{\,(procreate, HasFirstSubevent, find mate),}\textit{\,(procreate, Causes, have to raise your grandchild),}\textit{\,(procreate, HasFirstSubevent, form will to do so),}
  \item (go outside for even, MotivatedByGoal, see star): \textit{\,(go outside for even, HasSubevent, that you see star),}\textit{\,(go to film, UsedFor, see star),}\textit{\,(go outside for even, UsedFor, look at star),}\textit{\,(go outside for even, MotivatedByGoal, you have date),}\textit{\,(go outside for even, UsedFor, get out of house)}
 \end{itemize}\subsubsection{``same rel, similar word''}
        \begin{itemize}
        
  \item (aluminum, IsA, metal): \textit{\,(aluminum, IsA, material),}\textit{\,(safety-pins, MadeOf, metal),}\textit{\,(titanium, IsA, metal),}\textit{\,(quicksilver, IsA, metal),}\textit{\,(plumbum, IsA, metal),}
  \item (cherry, AtLocation, jar): \textit{\,(vegemite, AtLocation, jar),}\textit{\,(beet, AtLocation, jar),}\textit{\,(toffee, AtLocation, jar),}\textit{\,(jellybeans, AtLocation, jar),}\textit{\,(moonshine, AtLocation, jar),}
  \item (u.s president, IsA, political leader): \textit{\,(u.s president, IsA, in charge of arm force),}\textit{\,(president of something, IsA, it leader),}\textit{\,(president, IsA, leader),}\textit{\,(president, DefinedAs, leader of american government),}\textit{\,(us president, IsA, important political figure)}
 \end{itemize}\subsubsection{``different rel, similar word''}
        \begin{itemize}
        
  \item (attach case, AtLocation, embassy): \textit{\,(attach case, UsedFor, carry paper and book),}\textit{\,(attach case, AtLocation, office),}\textit{\,(attach case, AtLocation, courtroom),}\textit{\,(attache case, AtLocation, businessperson hand),}\textit{\,(attache case, CapableOf, hold important document),}
  \item (catch mumps, Causes, sickness): \textit{\,(die, HasSubevent, sickness),}\textit{\,(catch mumps, HasSubevent, you have fever),}\textit{\,(catch mumps, HasFirstSubevent, get sick),}\textit{\,(catch mumps, MotivatedByGoal, be sick),}\textit{\,(cold, IsA, sickness),}
  \item (buy something for love one, Causes, get lay): \textit{\,(get in line, MotivatedByGoal, get lay),}\textit{\,(have party, UsedFor, get lay),}\textit{\,(get pay, UsedFor, get lay),}\textit{\,(become inebriate, UsedFor, get lay)}
 \end{itemize}\subsubsection{``no directly related triple''}
        \begin{itemize}
        
  \item (fall from hot air balloon, CapableOf, kill you): \textit{\,(if you drink salt water it, CapableOf, kill you),}\textit{\,(drink sea water, CapableOf, kill you),}\textit{\,(water, CapableOf, kill you),}\textit{\,(lighten, CapableOf, kill you),}\textit{\,(pretty thing, CapableOf, kill you),}
  \item (milk, IsA, part of many food): \textit{\,(milk, DefinedAs, product of cow),}\textit{\,(milk, ReceivesAction, produce by female cow),}\textit{\,(milk, CapableOf, come from cow),}\textit{\,(milk, ReceivesAction, make into cheese),}\textit{\,(milk, ReceivesAction, create from cow),}
  \item (some food, ReceivesAction, make from dead animal): \textit{\,(some food, HasProperty, good but some be very dissgusting),}\textit{\,(some food, IsA, healthy and some be not),}\textit{\,(some food, HasProperty, poisonous if prepare improperly),}\textit{\,(some food, ReceivesAction, grind before eat),}\textit{\,(some food, HasProperty, consider exotic)}
 \end{itemize}
\subsection{Wikipedia}In this Section we report examples for each novelty category from Wikipedia dataset. For each example we include the $5$ triples that were shown to the human annotator, ordered by closeness according to our word embedding metric.\subsubsection{``same rel, rephrase''}
        \begin{itemize}
        
  \item (deep snow, IsA, winter): \textit{\,(snow, SymbolOf, winter),}\textit{\,(snow, AtLocation, winter),}\textit{\,(it, IsA, winter),}\textit{\,(snowflake, AtLocation, winter),}\textit{\,(nice time of year, IsA, winter time),}
  \item (winter season, HasProperty, cold): \textit{\,(winter weather, HasProperty, cold),}\textit{\,(in winter it, HasProperty, cold),}\textit{\,(snow fall from sky when weather, HasProperty, cold),}\textit{\,(stethascopes, HasProperty, cold),}\textit{\,(cold weather, Causes, cold),}
  \item (mathematical logic, HasProperty, logical): \textit{\,(mathmatics, HasProperty, logical),}\textit{\,(human wish for happiness but happiness, NotHasProperty, logical),}\textit{\,(design computer chip, HasPrerequisite, logical think),}\textit{\,(write program, HasPrerequisite, logical think),}\textit{\,(logic, DefinedAs, set of rule by which axiom can be manipulate to derive true statement)}
 \end{itemize}\subsubsection{``different rel, rephrase''}
        \begin{itemize}
        
  \item (the house, HasA, room): \textit{\,(house, MadeOf, room),}\textit{\,(many different way to put furniture, AtLocation, room),}\textit{\,(something you find upstairs, IsA, room),}\textit{\,(something you find downstairs, IsA, room),}\textit{\,(family room, IsA, room)}
 \end{itemize}\subsubsection{``same rel, similar word''}
        \begin{itemize}
        
  \item (bus system, AtLocation, city): \textit{\,(subway system, AtLocation, city),}\textit{\,(bus stop, AtLocation, city),}\textit{\,(bus, AtLocation, city),}\textit{\,(bus shelter, AtLocation, city),}\textit{\,(bus station, AtLocation, city),}
  \item (satellite radio, HasA, channel): \textit{\,(tv, HasA, channel),}\textit{\,(hear news, HasSubevent, change channel),}\textit{\,(watch television, HasSubevent, change channel),}\textit{\,(cnn, IsA, television channel),}\textit{\,(cnn, IsA, tv channel),}
  \item (summer, IsA, hotter weather): \textit{\,(summer, HasA, more sunshine than winter),}\textit{\,(summer, IsA, hot than winter),}\textit{\,(summer, IsA, warm than winter),}\textit{\,(summer, DefinedAs, season of baseball),}\textit{\,(summer, DefinedAs, warm season)}
 \end{itemize}\subsubsection{``different rel, similar word''}
        \begin{itemize}
        
  \item (liberal democracy, HasProperty, political):\textit{\,(democracy, IsA, political system),}\textit{\,(liberal democratic party, InstanceOf, japanese political party),}\textit{\,(feminism, IsA, political ideology),}\textit{\,(libertarianism, IsA, political ideology),}\textit{\,(liberalism, IsA, political ideology),}
  \item (music, UsedFor, musical express): \textit{\,(music, CapableOf, be express use musical notation),}\textit{\,(music, ReceivesAction, play with musical instrument),}\textit{\,(music, ReceivesAction, write with musical symbol),}\textit{\,(music, CreatedBy, instrument or human voice),}\textit{\,(music, CapableOf, express feel),}
  \item (the planet, HasA, mass): \textit{\,(boston, PartOf, mass),}\textit{\,(matter, HasA, mass),}\textit{\,(planet plutoi, ReceivesAction, discover by mr),}\textit{\,(some planet, HasA, more than one moon),}\textit{\,(magnitude of planet, IsA, quantifiable)}
 \end{itemize}\subsubsection{``no directly related triple''}
        \begin{itemize}
        
  \item (field, HasA, vector potential): \textit{\,(field, HasA, plant grow in them),}\textit{\,(field, UsedFor, agricultural pursuit),}\textit{\,(field, UsedFor, cultivate crop),}\textit{\,(field, UsedFor, graze livestock),}\textit{\,(field, UsedFor, ride horse),}
  \item (town, HasA, center of commerce): \textit{\,(town, ReceivesAction, compose of many neighborhood),}\textit{\,(town, HasProperty, likely to have several cafe),}\textit{\,(town, IsA, small than city),}\textit{\,(town, DefinedAs, prarie dog community),}\textit{\,(town, UsedFor, live in),}
  \item (divorce, HasProperty, mutual consent): \textit{\,(divorce, NotHasProperty, more common than marriage),}\textit{\,(divorce, DefinedAs, official end to marriage),}\textit{\,(divorce, IsA, fact of life),}\textit{\,(divorce, DefinedAs, termination of marriage),}\textit{\,(divorce, IsA, when marry couple separate legally)}
 \end{itemize}

\end{document}